\documentclass[conference]{IEEEtran}
\IEEEoverridecommandlockouts

\usepackage{cite}
\usepackage{amsmath,amssymb,amsfonts}
\usepackage{algorithmic}
\usepackage{graphicx}
\usepackage{textcomp}
\usepackage{xcolor}
\usepackage{booktabs}
\def\BibTeX{{\rm B\kern-.05em{\sc i\kern-.025em b}\kern-.08em
    T\kern-.1667em\lower.7ex\hbox{E}\kern-.125emX}}
\begin{document}

\title{A Resource-Efficient and Precision-Scalable FPGA Accelerator for Spiking Neural Networks Using Minifloat Arithmetic\\
\thanks{}
}

\author{\IEEEauthorblockN{Ali Mehrabi}
\IEEEauthorblockA{\textit{School of Engineering, Computing, and Science} \\
\textit{Western Sydney University}\\
Sydney, Australia \\
a.mehrabi@westernsydney.edu.au}
\and
\IEEEauthorblockN{Philip C. Jose}
\IEEEauthorblockA{\textit{School of Engineering, Computing, and Science} \\
\textit{Western Sydney University}\\
Sydney, Australia \\
p.jose@westernsydney.edu.au}

}

\maketitle

\begin{abstract}
Spiking neural networks (SNNs) are well suited to event-driven edge intelligence, but the arithmetic complexity, memory footprint, and power consumption associated with conventional floating-point implementations can limit their deployment on resource-constrained hardware. This paper presents a precision-scalable and resource-efficient SNN hardware accelerator architecture based on parameterized minifloat arithmetic. The proposed architecture implements recurrent leaky integrate-and-fire (LIF) neurons and leaky-integrator (LI) readout neurons using a unified minifloat datapath whose word length and mantissa width can be configured at synthesis time. A valid-driven arithmetic subsystem employs pipelined minifloat multiplication and addition, while time-multiplexed processing enables arithmetic resources to be shared across synaptic accumulation, neuron-state updates, recurrent connections, and output processing. This organization reduces replicated arithmetic hardware and provides a scalable design approach for different SNN sizes and precision requirements. The architecture supports E8M23, E5M10, E5M6, and E4M3 formats within the same RTL framework, allowing precision, memory capacity, timing, and power to be traded without modifying the underlying neuron architecture. FPGA implementation results demonstrate that reducing minifloat precision substantially decreases arithmetic-resource utilization and estimated dynamic power, while also reducing the theoretical storage requirement of synaptic parameters. A recurrent event-based classification network is used as a hardware validation case, demonstrating that substantial precision reduction can be achieved while retaining competitive inference accuracy. The results establish parameterized minifloat computation as an effective basis for scalable, low-resource, and energy-efficient FPGA SNN accelerators.
\end{abstract}

\begin{IEEEkeywords}
Spiking Neural Networks, Minifloat, Leaky Integrate-and-Fire, FPGA, Neuromorphic Computing.
\end{IEEEkeywords}

\section{Introduction}
Spiking neural networks (SNNs) combine discrete communication events with temporal neuron state and are attractive for low-latency, event-driven edge intelligence. Unlike conventional artificial neural networks, an SNN retains dynamical state through variables such as membrane potential and synaptic current, allowing information to be represented by the timing and sparsity of spikes \cite{maass1997,gerstner2014}. Surrogate-gradient training has made recurrent SNNs practical to train while preserving event-driven inference behavior \cite{neftci2019}. These properties motivate dedicated hardware that can exploit temporal sparsity, deterministic execution, and local state updates.

Field-programmable gate arrays (FPGAs) are particularly suitable for SNN acceleration because they provide configurable parallelism, distributed state storage, deeply pipelined datapaths, deterministic timing, and rapid adaptation of the neuron and synapse architecture. However, numerical representation remains a major hardware design decision. Although spikes themselves are binary, synaptic weights, membrane potentials, currents, thresholds, and decay coefficients are usually multi-bit quantities. Full IEEE~754 single-precision arithmetic offers large dynamic range and familiar software behavior but incurs considerable logic, register, routing, and power overhead. Fixed-point arithmetic can reduce this cost, but selecting a fixed scaling factor becomes difficult when weights and neuron states span different numerical ranges.

Reduced-precision floating-point arithmetic, or minifloat, provides an attractive design space for SNN hardware because the exponent preserves a useful dynamic range while the mantissa and total word length can be reduced to lower memory, arithmetic, routing, and switching costs. In this work, numerical precision is treated as a hardware design parameter rather than as a fixed property of the network. Depending on the accuracy requirements and resource constraints of a target application, the minifloat word length can therefore be selected to trade numerical precision against silicon area, FPGA resource utilization, dynamic power, and operating frequency. The objective is to provide a single SNN processor architecture that can be scaled across these operating points without redesigning the neuron model, event-processing flow, or recurrent-network control structure.

To achieve this objective, a parameterized SNN accelerator is proposed in which event handling, synaptic accumulation, neuron-state computation, recurrent-spike management, readout processing, memory access, and global control are organized as reusable hardware subsystems. The accelerator employs a shared minifloat arithmetic datapath parameterized by $(P_{\mathrm{BITS}},P_{\mathrm{MANTISSA}})$, with
$
E=P_{\mathrm{BITS}}-P_{\mathrm{MANTISSA}}-1,
$
so that the exponent and mantissa widths can be changed at synthesis time. Format-dependent neuron constants are selected internally by the RTL, allowing the same architecture and scheduler to support E8M23, E5M10, E5M6, and E4M3 implementations. To make the accelerator suitable for area- and power-constrained FPGA implementations, its processing architecture is time-multiplexed, allowing the same arithmetic resources to be reused for synaptic accumulation, recurrent processing, LIF neuron updates, and output-layer computation.

The main contributions of this work are as follows:
\begin{enumerate}
    \item A scalable and resource-efficient FPGA SNN processor architecture in which event processing, synaptic accumulation, recurrent communication, LIF/LI neuron updates, output readout, memory, and control are organized as reusable hardware blocks. The architecture is independent of a specific application and can be configured for different network dimensions and numerical precision requirements.

    \item A parameterized minifloat LIF/LI neuron datapath in which numerical precision is controlled through $P_{\mathrm{BITS}}$ and $P_{\mathrm{MANTISSA}}$. This allows the same RTL architecture to support multiple floating-point formats, including E8M23, E5M10, E5M6, and E4M3, without redesigning the neuron-processing hardware.

    \item A hardware architecture that exposes an explicit application-dependent trade-off between numerical precision and implementation cost. The minifloat word length can be selected according to the required inference accuracy, available FPGA resources, power budget, and target operating frequency, enabling the accelerator to be scaled from higher-precision implementations to compact low-power realizations.

    \item A valid-driven and deeply pipelined minifloat arithmetic subsystem comprising a six-stage adder and a four-stage multiplier. These arithmetic units are time-multiplexed and reused across synaptic accumulation and neuron-state equations, minimizing hardware duplication while maintaining deterministic processing.

    \item A post-route FPGA design-space evaluation on an XCU250 device demonstrating the implementation benefits of precision scaling. Relative to the E8M23 implementation, the E5M6 configuration reduces CLB, LUT, flip-flop, and estimated dynamic-power requirements by 69.2\%, 54.4\%, 59.1\%, and 46.0\%, respectively, while increasing the estimated maximum operating frequency by 18.5\%.

    \item Hardware validation using an event-driven recurrent SNN workload, demonstrating that substantial reductions in arithmetic precision and hardware cost can be achieved while retaining the same classification accuracy as the higher-precision RTL baseline over the tested E5M10 and E5M6 configurations. The application experiment is used as a validation of the proposed accelerator architecture rather than as the primary contribution of the work.
\end{enumerate}

\section{Background and Related Work}
\subsection{FPGA-Based SNN Hardware}
Dedicated neuromorphic processors such as TrueNorth and Loihi demonstrate the benefits of state-local, event-driven neural computation \cite{merolla2014,davies2018}. FPGAs provide a complementary platform in which neuron models, synaptic storage, communication, and arithmetic can be customized to the target workload. A recent survey identifies neuron realization, memory organization, resource sharing, network scalability, and numerical representation as major design dimensions in FPGA SNN accelerators \cite{karamimanesh2025}.

The fully connected, multi-layer, gradient-free online-trainable SNN architecture reported in \cite{mehrabi2024access, Philip} demonstrates that learning and inference can be integrated within a single FPGA implementation. However, the hardware and memory requirements of a fully connected realization increase significantly with network size, limiting its suitability for large-scale SNN deployment. In contrast, scalable FPGA SNN accelerators commonly rely on time-multiplexed processing, shared arithmetic resources, and centralized or distributed state memories to reduce hardware replication and improve area efficiency \cite{mehrabi2024chapter}. 

\subsection{Reduced-Precision Floating Point for Neural Hardware}
IEEE~754 defines general floating-point arithmetic and special-value behavior \cite{ieee754}. Neural workloads, however, often tolerate reduced mantissa precision. Mixed-precision and FP8 studies have shown that floating-point word length can be reduced substantially when numerical range and scaling are appropriately managed \cite{micikevicius2018,wang2018fp8,sun2019hfp8,micikevicius2022fp8}. For SNNs, Wang \emph{et al.} investigated reduced-precision memory and computation and reported favorable FPGA efficiency for low-bit floating-point representations \cite{wang2019snn}. Quantized SNN studies have also shown that weight and membrane precision can be reduced aggressively, although the tolerable precision depends on network dynamics and training strategy \cite{schaefer2023,hassan2024}. An adaptive FPGA LIF implementation further illustrates the direct impact of neuron-datapath simplification on speed and energy \cite{mishra2025}.

The present work differs in emphasis: the contribution is a unified, parameterized hardware architecture and its post-route precision scaling, rather than a new quantization or training algorithm. The same control flow and neuron equations are retained while the floating-point datapath width is varied.

\section{Precision-Scalable Accelerator Architecture}
\subsection{System-Level Organization}
\begin{figure*}[t]
    \centering
    \includegraphics[width=0.95\textwidth]{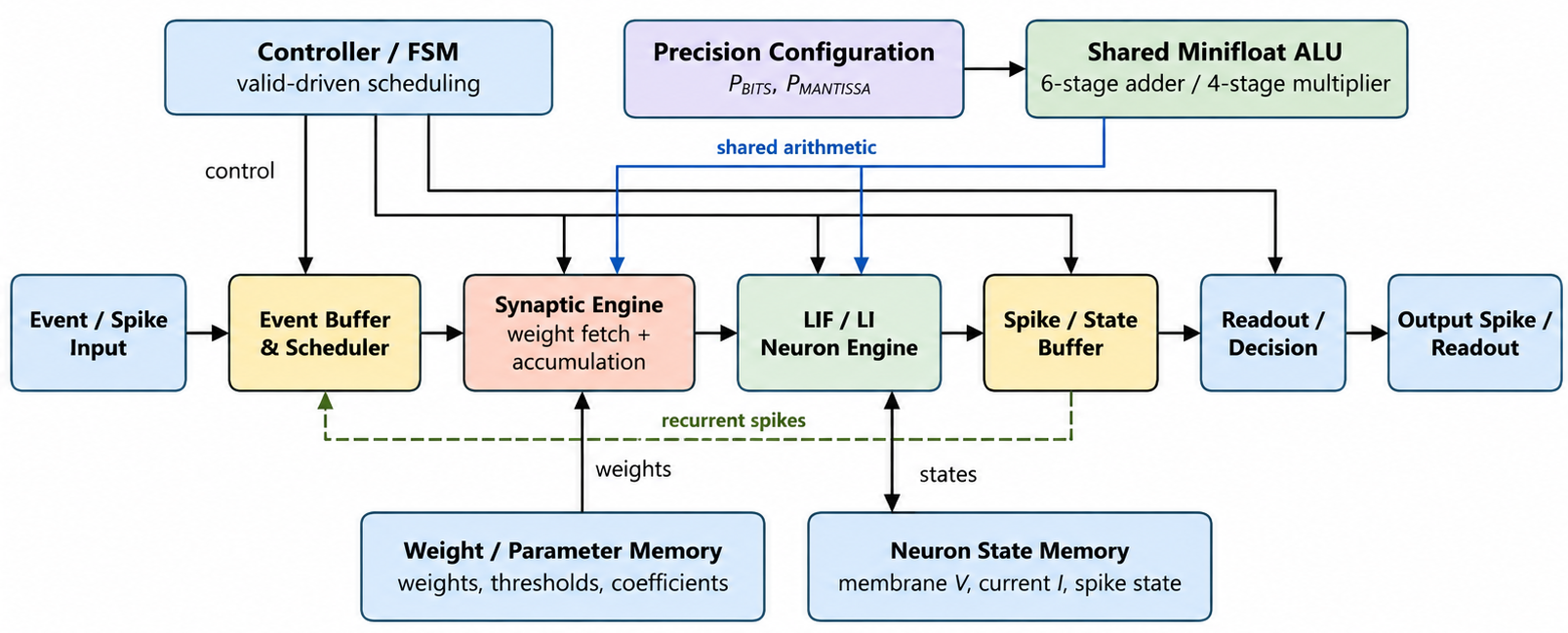}
    \caption{Generic architecture of the precision-scalable SNN accelerator. Network topology, parameters, and numerical precision are configurable.}
    \label{fig:architecture}
\end{figure*}

Fig.~\ref{fig:architecture} shows the general architecture of the proposed accelerator. The datapath is independent of any particular sensor or dataset. Events or spikes enter through an input interface and are queued by an event buffer and scheduler. The synaptic engine fetches the required weights and accumulates input and recurrent contributions. The resulting synaptic current is passed to a time-multiplexed LIF/LI neuron engine. Generated spikes and updated states are written to the spike/state buffer and state memory; recurrent spikes are fed back to the synaptic engine, while output activity is passed to a readout stage.

A controller/FSM orchestrates all operations. Rather than assuming combinational or single-cycle floating-point arithmetic, the scheduler launches an arithmetic operation with an input-valid pulse, waits for the corresponding output-valid indication, and only then advances the neuron or synapse state machine. This valid-driven execution model is essential because the adder and multiplier have different pipeline depths.

The architecture logically separates parameter memory from neuron-state memory. Parameter memory contains synaptic weights, thresholds, and constant coefficients, whereas state memory contains membrane potential, synaptic current, and spike state. This separation permits alternative mappings: parameters can reside in on-chip memories for smaller networks or in external memory for larger networks, while frequently updated neuron state can remain close to the compute engine. The post-route results reported in Section~VI characterize the measured compute core; the current synthesized design does not infer BRAM or URAM for the benchmark weight arrays, so learned-weight capacity is reported separately as a raw storage requirement.

\subsection{Time-Multiplexed Processing}
The architecture uses time multiplexing to decouple the neuron count from the number of instantiated floating-point operators. A synaptic accumulator is shared across active input events and recurrent spikes. Hidden neurons are then updated sequentially by a common LIF arithmetic engine, and output neurons are evaluated by a shared LI engine. This organization deliberately favors a clear precision-scaling baseline over maximum spatial parallelism. More processing elements can be replicated in future implementations without changing the minifloat format or local neuron equations.

For a recurrent hidden layer, the accumulated synaptic current can be written generally as
\begin{equation}
I_{\mathrm{syn},j}[t] = \sum_i W^{\mathrm{in}}_{ji}x_i[t] + \sum_k W^{\mathrm{rec}}_{jk}z_k[t-1],
\end{equation}
where $x_i[t]$ denotes the input event/spike activity and $z_k[t-1]$ denotes the previous hidden spike state. This interface allows the same neuron engine to be reused for different input modalities and network sizes.

\subsection{Precision Configuration}
The minifloat word length is defined by
\begin{equation}
N = 1 + E + M,
\end{equation}
where one bit is used for sign, $E$ for exponent, and $M$ for the stored fraction. The exponent width is derived as $E=N-M-1$. For normalized finite values,
\begin{equation}
x=(-1)^s\left(1+\frac{F}{2^M}\right)2^{e-B},\qquad B=2^{E-1}-1.
\end{equation}
The custom arithmetic is IEEE-like rather than fully IEEE~754 compliant: the implementation is optimized for SNN inference and does not require the complete NaN/infinity/subnormal machinery of a general-purpose floating-point unit.

\begin{table}[t]
\caption{Evaluated Numerical Formats}
\label{tab:formats}
\centering
\begin{tabular}{lcccc}
\toprule
Format & $N$ & Sign & $E$ & $M$ \\
\midrule
E8M23  & 32 & 1 & 8 & 23 \\
E5M10  & 16 & 1 & 5 & 10 \\
E5M6   & 12 & 1 & 5 & 6 \\
E4M3   & 8  & 1 & 4 & 3 \\
\bottomrule
\end{tabular}
\end{table}

\section{LIF Neuron and Minifloat Arithmetic}
\subsection{Exact Neuron State Update}
\begin{figure*}[t]
    \centering
    \includegraphics[width=0.95\textwidth]{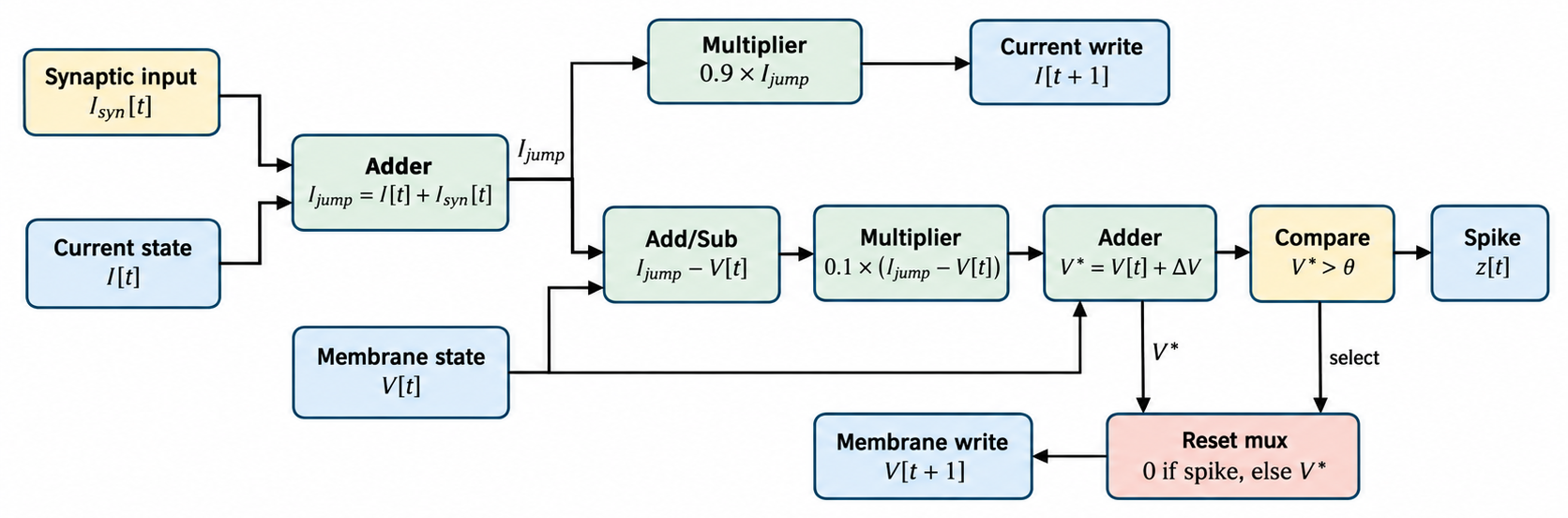}
    \caption{LIF neuron datapath with parameterized minifloat arithmetic. The block sequence follows the RTL update order, and the shared adder and multiplier are realized as valid-driven pipelined arithmetic units.}
    \label{fig:lif}
\end{figure*}

Fig.~\ref{fig:lif} shows the implemented LIF datapath together with the parameterized arithmetic subsystem. The hidden neuron first forms a current jump,
\begin{equation}
I_{\mathrm{jump}}[t]=I[t]+I_{\mathrm{syn}}[t].
\end{equation}
The membrane candidate is then updated according to
\begin{equation}
V^{*}[t]=V[t]+k_m\left(I_{\mathrm{jump}}[t]-V[t]\right),
\end{equation}
with $k_m=0.1$ in the evaluated model. The synaptic-current state decays as
\begin{equation}
I[t+1]=\alpha I_{\mathrm{jump}}[t],\qquad \alpha=0.9.
\end{equation}
A spike is generated when $V^{*}[t]>\theta$. The next membrane state is zero after a spike and otherwise retains $V^{*}[t]$.

The output readout uses an LI cell with the same membrane integration coefficient $k_m=0.1$ but without threshold/reset. Its current decay coefficient is $0.8$. Keeping hidden LIF and output LI dynamics in the same minifloat domain avoids repeated format conversion and permits arithmetic reuse.

\subsection{Pipelined Minifloat Adder and Multiplier}
The minifloat multiplier uses four pipeline stages: field unpacking, mantissa multiplication with exponent arithmetic, normalization, and result packing. The adder uses six stages: field unpacking, exponent comparison/alignment, mantissa addition or subtraction, normalization, fraction truncation, and result packing. Zero handling is explicitly implemented so that an all-zero operand does not acquire an implicit leading one.

These pipeline depths are deliberately preserved across all tested precisions. Consequently, reducing precision changes the width and cost of the arithmetic datapath without changing the high-level scheduler or number of arithmetic transactions. This makes post-route comparisons more meaningful because precision is the principal architectural variable.

\section{Implementation and Validation Methodology}
\subsection{RTL and FPGA Flow}
The accelerator is described in synthesizable Verilog and parameterized by total word length and mantissa width. Format-dependent constants, including $0.1$, $0.9$, $0.8$, and the threshold, are selected internally from $(P_{\mathrm{BITS}},P_{\mathrm{MANTISSA}})$ so that vector constants do not need to be passed through synthesis Tcl. Each precision is synthesized, optimized, placed, physically optimized, routed, and physically optimized. 

All four implementations target the AMD Xilinx XCU250 device (part {xcu250-figd2104-2l-e}). Estimated maximum frequency is calculated from the final setup slack and the applied clock constraint as
\begin{equation}
F_{\max}\approx \frac{1000}{T_{\mathrm{constraint}}-\mathrm{WNS}}\quad \text{MHz},
\end{equation}
 where, WNS is the worst negative slack.
 
The Python reference model, parameterized Verilog RTL, testbench, FPGA constraints, and Vivado Tcl scripts used in this work are publicly available in the project repository at \cite{mehrabi_mf_snn}. The repository provides the files required to reproduce the software-to-RTL validation and FPGA implementation flow.

\subsection{Controlled Event-Based Validation}
To verify numerical consistency between the software model and RTL implementation, a recurrent event-based classifier is used as a compact validation workload. The network comprises 2,450 input channels, 100 recurrent hidden LIF neurons, and four LI readout neurons. A Python model is trained once, and the same network topology, temporal event bins, trained weights, and inference schedule are retained across all RTL precision configurations. The 131-recording Poker-DVS dataset \cite{serrano2015} is divided deterministically into 104 training and 27 test recordings, with events accumulated into 1-ms bins before inference. Norse \cite{pehle2021} is used to implement the software SNN reference model.

The benchmark is used only to test whether numerical precision changes the recurrent spike trajectory and final decision. It is not intended as a state-of-the-art event-recognition study.

\section{Results and Discussion}
\subsection{Hardware Scaling}
Table~\ref{tab:hardware} summarizes the final routed results. Reducing precision produces a strong and mostly monotonic reduction in logic and register usage. E5M10 reduces CLBs from 2,158 to 776 and LUTs from 3,990 to 2,253. E5M6 further reduces the design to 665 CLBs, 1,818 LUTs, and 3,717 flip-flops. E4M3 is the smallest configuration at 464 CLBs, 1,339 LUTs, and 2,618 flip-flops.

\begin{table*}[t]
\caption{Post-Route FPGA Comparison on XCU250}
\label{tab:hardware}
\centering
\begin{tabular}{lrrrrrrrr}
\toprule
Format & $F_{\max}$ (MHz) & WNS (ns) & CLBs & LUTs & FFs & DSPs & BRAM/URAM & Dynamic Power (W) \\
\midrule
E8M23 & 226.603 & 0.587 & 2158 & 3990 & 9094 & 4 & 0/0 & 0.298 \\
E5M10 & 256.937 & 1.108 & 776  & 2253 & 4819 & 2 & 0/0 & 0.196 \\
E5M6  & 268.528 & 1.276 & 665  & 1818 & 3717 & 2 & 0/0 & 0.161 \\
E4M3  & 262.605 & 1.192 & 464  & 1339 & 2618 & 2 & 0/0 & 0.147 \\
\bottomrule
\end{tabular}
\end{table*}

Relative to E8M23, E5M6 reduces CLBs by 69.2\%, LUTs by 54.4\%, flip-flops by 59.1\%, and estimated dynamic power by 46.0\%. At the same time, estimated $F_{\max}$ increases from 226.603 to 268.528~MHz, an 18.5\% improvement. E4M3 provides the greatest logic and power reduction, with 78.5\% fewer CLBs, 66.4\% fewer LUTs, 71.2\% fewer flip-flops, and 50.7\% lower dynamic power than E8M23. Its $F_{\max}$ is slightly below E5M6 but remains 15.9\% above the E8M23 implementation. The DSP count falls from four in E8M23 to two in each reduced-precision design.

The routed reports list zero BRAM and URAM for all four variants. The reported hardware comparison therefore characterizes the arithmetic/control/state core generated by the present RTL flow rather than a fully integrated BRAM-resident synaptic-memory subsystem. This distinction is important: the logical accelerator architecture in Fig.~\ref{fig:architecture} includes parameter memory, but the current benchmark weights are not represented as inferred on-chip RAM in the measured implementation.

\subsection{Accuracy and Weight-Storage Trade-Off}
The Python reference model achieved 100\% classification accuracy The corresponding hardware accuracies for the evaluated minifloat formats are summarized in Table~\ref{tab:accuracy}.
The E8M23, E5M10, and E5M6 RTL implementations each correctly classified with 96.30\% accuracy.
The E4M3 implementation achieved an accuracy of 88.89\%. These results show that reducing
the precision from E8M23 to E5M6 introduces no additional classification error relative to the higher-precision RTL baseline, while E4M3 begins to
degrade the recurrent network behavior. Table~\ref{tab:accuracy} also
reports the raw storage required by the 255,400 learned weights for each
numerical format.

\begin{table}[t]
\caption{Validation Accuracy and Raw Learned-Weight Storage}
\label{tab:accuracy}
\centering
\begin{tabular}{lccc}
\toprule
Format & RTL Accuracy & Storage (KiB) & Reduction \\
\midrule
E8M23 & 96.30\% & 997.7 & baseline \\
E5M10 & 96.30\% & 498.8 & 50.0\% \\
E5M6  & 96.30\% & 374.1 & 62.5\% \\
E4M3  & 88.89\% & 249.4 & 75.0\% \\
\bottomrule
\end{tabular}
\end{table}

E5M6 is therefore the strongest measured operating point for this implementation: it preserves the E8M23 RTL accuracy while delivering the highest $F_{\max}$ and large reductions in logic, registers, power, and theoretical learned-weight storage. E4M3 demonstrates the next point in the trade-off curve: hardware cost continues to fall, but classification accuracy decreases by 7.41 percentage points relative to the E8M23 RTL baseline.

\subsection{Architectural Implications and Limitations}
The results support three architectural observations. First, mantissa reduction can provide substantial logic and register savings even when the overall controller and neuron schedule remain unchanged. Second, the non-monotonic $F_{\max}$ behavior between E5M6 and E4M3 shows that post-route timing depends on placement, routing, and critical-path structure in addition to datapath width; the smallest format is not automatically the fastest. Third, precision must be selected jointly with neuron dynamics. E4M3 is highly efficient in hardware, but its reduced state resolution alters enough recurrent behavior to change final classifications on the validation workload.

\section{Conclusion}
This paper presented a general precision-scalable minifloat FPGA accelerator architecture for SNNs. The design separates event scheduling, synaptic accumulation, recurrent spike handling, neuron-state memory, readout, and valid-driven arithmetic so that numerical precision can be changed without modifying the system-level control architecture. A six-stage adder and four-stage multiplier are shared by time-multiplexed LIF/LI neuron and synaptic-processing operations, and the RTL supports E8M23, E5M10, E5M6, and E4M3 through two format parameters.

Post-route results on an XCU250 show that E5M6 provides the strongest measured trade-off: compared with E8M23 it reduces CLB usage by 69.2\%, LUTs by 54.4\%, flip-flops by 59.1\%, and estimated dynamic power by 46.0\%, while increasing estimated $F_{\max}$ by 18.5\%. On the controlled recurrent event-based validation workload, E5M6 also retains the same 96.30\% RTL accuracy as the 32-bit baseline. E4M3 further reduces hardware cost but decreases accuracy to 88.89\%. These results demonstrate that parameterized minifloat arithmetic can be integrated as an architectural design-space dimension in FPGA SNN accelerators rather than treated only as an application-specific quantization choice.

\bibliographystyle{IEEEtran}
\bibliography{Bibliography}

\end{document}